\documentclass{bvm} % Do NOT change this line

\begin{document}
% Do not change anything in the preamble (anything above \begin{document}) except for the specification of the bibliography file, any additional changes will be lost

% You can create your own commands using \newcommand. These must be placed after(!) \begin{document} and should contain your own name to avoid multiple definitions between different papers
\newcommand{\bvmyear}{2026}

% use the \selectlanguage command to select the language in which your proceedings are written
%\selectlanguage{ngerman} % German
\selectlanguage{english} % English

% The title of your paper
\title{Fracture Morphology Classification}
% If you write a short paper/abstract, the title must start with "Abstract:".
% \title{Abstract: Bildverarbeitung für die Medizin \bvmyear}

% Optional specification of a subtitle
\subtitle{Local Multiclass Modeling for Multilabel Complexity}

% titlerunning appears in the header of every second page
% LaTeX generates this automatically from your paper title
% However, if it is too long, the message "Title Suppressed Due to Excessive Length" appears instead.
% In this case, specify an abbreviated form of the title here
\titlerunning{Fracture Morphology Classification}

% Please indicate all authors involved
% To allow us to correctly identify the last name of each author, indicate it using the \lname{} command.
% If you want to provide your orcid number, you can use the optional argument of \lname (\lname[your_orcid_number]{your_last_name}).
% If more than one institute is involved, list the number of the institute(s) (see below) with \inst{} after the respective author. If only one institute is involved, omit this.
% Separate all authors with a comma
% Use the command `\equalContribution' for shared main authorship after the institutes of the corresponding authors, e.g. `\fname{Thomas~M.} \lname{Deserno} \inst{1}\equalContribution, \fanem{Heinz} \lname{Handels} \inst{2}\equalContribution,`.
% Provide the extra information for each author according to the guidelines. The publisher requires these information.
\author{
	\fname{Cassandra} \lname{Krause} \affiliation{Universität zu Lübeck} \authorsEmail{cassandra.krause@student.uni-luebeck.de} \street{Ratzeburger Allee} \housenumber{160} \zipcode{23562} \city{Lübeck} \country{Germany} \isResponsibleAuthor,
	\fname{Mattias P.} \lname[0000-0002-3499-4328]{Heinrich}\affiliation{Universität zu Lübeck} \authorsEmail{mattias.heinrich@uni-luebeck.de} \street{Ratzeburger Allee} \housenumber{160} \zipcode{23562} \city{Lübeck} \country{Germany},
	\fname{Ron} \lname[0009-0006-7583-9144]{Keuth} \affiliation{Universität zu Lübeck} \authorsEmail{ron.keuth@uni-luebeck.de} \street{Ratzeburger Allee} \housenumber{160} \zipcode{23562} \city{Lübeck} \country{Germany}
}

% Enter the authors here as you want them to appear in the header
% Name only the surnames
% Depending on the number of authors involved, follow the examples below
% \authorrunning{Meier} - one author
% \authorrunning{Meier \& Müller} - two authors
% \authorrunning{Meier, Müller \& Schulze} - three authors
% \authorrunning{Meier et al.} - more than three authors
\authorrunning{Krause, Heinrich \& Keuth}

% Specify the institutes involved
% In case of participation of more than one institute, each institute shall be preceded by an ascending number with \inst{}.
% If only one institute is involved, omit the corresponding number.
% Separate individual institutes with \\
\institute{
Institut für Medizinische Informatik, Universität zu Lübeck
}

% Enter the e-mail address of the corresponding author
\email{ }

\maketitle

% Abstract of your paper, only for long papers
% Do NOT use \begin{abstract} ... \end{abstract} for short articles
\begin{abstract}
	Between $15\,\%$ and $45\,\%$ of children experience a fracture during their growth years, making accurate diagnosis essential. Fracture morphology, alongside location and fragment angle, is a key diagnostic feature. In this work, we propose a method to extract fracture morphology by assigning global AO codes to corresponding fracture bounding boxes. This approach enables the use of public datasets and reformulates the global multilabel task into a local multiclass one, improving the average F1 score by $7.89\,\%$. However, performance declines when using imperfect fracture detectors, highlighting challenges for real-world deployment. Our code is available on GitHub.
\end{abstract}

\section{Introduction}
$15\,\%$ to $45\,\%$ of all children suffer a fracture by the end of their growing years \cite{Kraus2005}, with the distal forearm being the most common location.
Consequently, proper treatment of wrist bones fractures is particularly important in the paediatric field to avoid permanent deformities or growth disorders.
Fracture morphology, alongside location and fragment angle, is a key characteristic for describing fractures \cite{Khan2023}.
Hence, a reliable morphology classification is crucial for therapy planning.
The AO/OTA system (Arbeitsgemeinschaft für Osteosynthesefragen/Orthopaedic Trauma Association) offers a standardized framework for classifying fractures by location and morphology \cite{Slongo2006}.
Based on this system, several deep learning-based approaches for binary fracture classification have emerged lately, ranging from supervised \cite{Shojib2024} to self-supervised methods \cite{Thorat2024}.
For a finer classification, other approaches include the location of fractures covering radius/ulna and their metaphyse/epiphyses \cite{Binh2024} or identify extra-, partial-, and intra-articular fractures\cite{Min2023, Gan2024, Yany2021}.
Recent work extends the input modality to segmentation and radiology reports to classify the seven most common AO classes.
Facing challenges, they conclude that a hierarchical stage could be better suited for AO/OTA systems, as it was done by \cite{Min2023, Gan2024}.
Both first detecting ROIs and then classifying fractures within two separated models.
To our best knowledge, we provide the first study explicitly focusing on fracture morphology.
With this, we take the first step to a novel approach for fracture classification within the AO system, where we aim to provide the different code components separately, with this work covering classification of fracture morphology.

\section{Materials and methods}
\subsection{Dataset}
We use the public available GRAZPEDWRI-DX \cite{Nagy2022} dataset.
It holds $20\,327$ paediatric trauma wrist X-ray images of AP and lateral view combined with bounding boxes and AO codes for $18\,090$ fractures. 
% including diaphyseal (22-D), distal metaphysial (23-M) and distal epiphyseal (23-E) fractures \cite{Slongo2006}.
For preprocessing, we follow \cite{Nagy2022} and relative split the dataset into 8:1:1 train/validation/test images, preserving the label distribution.
%(https://doi.org/10.6084/m9.figshare.19330688.v1)

\subsection{Extraction of fracture morphology from AO codes}\label{sec:morphology-classification}
\begin{figure}[hb]
    \centering
    \includegraphics[width=1\linewidth]{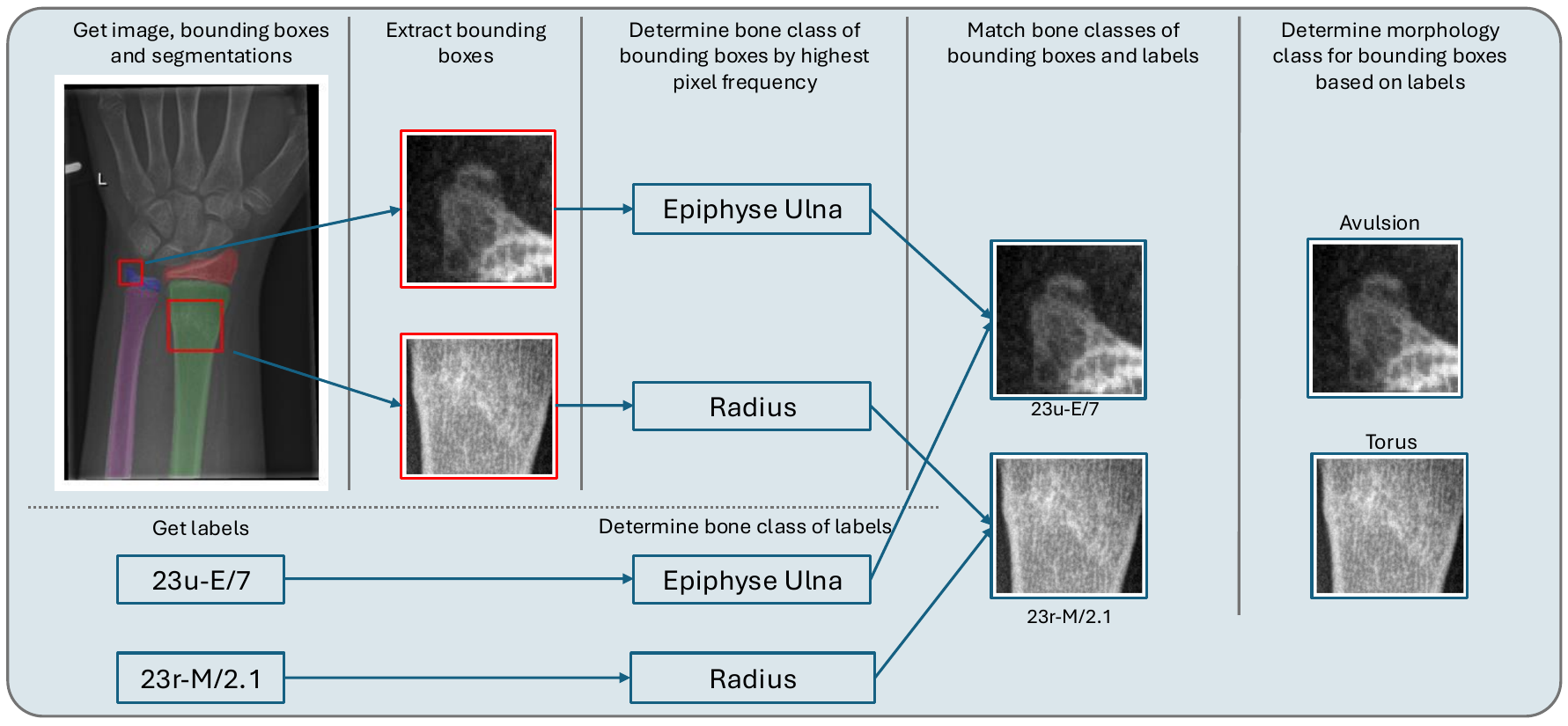}
    \caption{Proposed pipeline to extract fracture morphology by assigning global AO codes to the corresponding fracture bounding boxes. Therefore, a matching of the bounding boxes and labels is performed based on their corresponding bone class.}
    \label{fig:pipeline}
    \altText{Pipeline for the fracture morphology classification. First the X-ra image, the bounding boxes and segmentations are extracted. From this, the bounding boxes are extracted and the bone class is determined for each bounding box based in the highest pixel frequency. In parallel, the labels for the image are extracted and also for them the bone class is determined. Then, the bone classes of the bounding boxes and labels are marched and finally for each the morphology class is determined.}
\end{figure}

Since the dataset itself does not provide any fracture morphology labels natively, we have to extract them from the provided AO-codes and fractures' bounding boxes.
We rely on the classification of Kaiser and Weinberg  for femoral shaft fractures and Salter classification (I-IV) regarding epiphyseal fractures.
With this, we create a mapping from the provided global AO codes to their corresponding fracture morphology in collaboration with radiologists.
While mapping a single fracture to its AO code is straightforward (1:1), associating multiple fractures with their respective codes is ambiguous, as AO codes are assigned globally to each image.
To resolve this, we use the segmentation masks from \cite{Keuth2025DenseSeg} for the radius, ulna, and their epiphyses, assigning each fracture to the bone with the greatest segmentation overlap within its bounding box (Fig.~\ref{fig:pipeline}) and thus extract its bone label.
Simultaneously, AO codes are assigned to the respective bone labels using a further mapping from AO codes to bone classes.
If an AO code refers to fractures occurring in both the radius and the ulna (e.g. 22-D/4.1), it is replaced by their two variants (22r-D/4.1 and 22u-D/4.1 respectively), since two bounding boxes are given in these cases.
With this, we can map each fracture (bounding box) to a AO-code and vice versa and assign its fracture morphology label (Fig.~\ref{fig:pipeline}).
We find that six of eleven fracture morphology have too few samples to train our models, even after applying loss weighting and oversampling.
Consequently, we exclude those from our experiments.
Fig.~\ref{fig:f1perclass} shows the five fracture morphologies extracted from the 18 different AO-codes.
%Additionally, we referred to false-positives as “Healthy”.
%We determine a false-positive as a patch without a $\text{IoU} < 0.5$ to any ground truth bounding box. 

\subsection{Experiments}
As a \emph{lower baseline}, we train a CNN on the whole image utilizing morphology classes provided by the global AO-codes.
With this, no mapping from morphology class to the fractures exits and hence we only detect their global presence, formulating this task as a \emph{multilabel classification} (multi-label on full-image).
For our \emph{multiclass approach}, we extract the fracture's bounding boxes as patches and assign their morphology label utilizing our pipeline described in Sec.~\ref{sec:morphology-classification}.
On those patches, we train a CNN converting the fracture morphology classification from a multi-label full-image into a multi-class patch-based task.
However, extracting those patched requires access to the fractures' bounding boxes and while we use the provided ground truth (GT) bounding boxes as an \emph{upper baseline} (multi-class on GT BBox) they are not available in a real-world application with unknown images.
To overcome this limitation, we also consider the impact of a non-perfect fracture detection.
We train a YOLO detector on the same dataset with the GT fracture bounding boxes.
Here, we experiment with different confidence thresholds $t\in\{0.01, 0.05, 0.1, 0.5, 0.8, 0.85\}$ for YOLO's fracture detection to further boost its recall, since for a real-world application, an overlooked fracture would be a crucial error.
Since a low $t$ results in a higher recall but lower precision, we implement a \emph{false-positive reduction} (FP-reduction).
For this, we extend our classifier with a new “Healthy” class holding all patches proposed by YOLO covering non-fracture image content.
We determine such a false-positive by comparing YOLO's predicted bounding box with the GT.
If no GT box exist having an IoU of at least 0.5 (threshold of the Pascal VOC challenge), we consider this prediction as false-positive and assign the label “Healthy”.

\subsection{Training setup}
As CNN for classification, we chose an ImageNet-pretrained ResNet18 (\texttt{torchvision} implementation).
We employ a drop out layer ($p=0.1$) before the classifier head to lessen overfitting.
For the patch-based approach, we remove the stride of the first conv layer with kernel size 7 and its following max pooling to adapt the model's downsample scale to the lower patch resolution.
As preprocessing for the full-image baseline, we bilinear resize the images to $384\times224$ preserving the average aspect ratio of the dataset and providing enough details to detect fine fracture lines.
The patches were extracted from the original high-resolution image and bilinear resized to their average size ($96\times96$).
% All inputs a $z$-standardized with the statistic of the corresponding training split and expanded to three channel to ensure compatibility with the pretrained weights.
We counter class imbalance by employing loss weighting and oversampling via a weighted random sample using the class inverse frequencies as weights.
For on-the-fly data augmentation, we use random affine transformations: rotation of $\mathcal{U}(-25^\circ, 25^\circ)$, scaling of $\mathcal{U}(80\,\%, 120\,\%)$, and translation of $\mathcal{U}(-10\,\%, 10\,\%)$ in width and height.
The Adam optimizer minimizes the loss (cross entropy for multiclass and binary cross entropy for mutlilabel) for 200 epochs.
We adapt its initial learning rate $1e^{-4}$ using a cosine annealing scheduler with a single cycle and a linear warm-up during the first 20 epochs.
For fracture detection, we employ a YOLOv10x (\texttt{ultralytics} framework) and train it with a three folded cross validation.
For our morphology classification, we then use the model that included the current image in its validation split.
We evaluate our experiments with the macro-averaged accuracy, F1-score, precision, and recall.
Fractures missed by YOLO (false negatives) are considered errors and are included in the evaluation metrics.
Our code is available on GitHub\footnote{\url{https://github.com/multimodallearning/FractureMorphologyClassification}}.%\todo{add repo}

\section{Results}
The quantitative results in Tab.~\ref{tab:comparison method} show both baselines outperforming the fracture morphology classification employing a YOLO-based fracture detection.
The multi-label approach, taking the whole image as input, yields the highest accuracy.
When the GT fracture bounding boxes are used in our patch-based, multi-class approach, we achieve the highest recall, precision and thus F1-Score.
However, when we use the YOLO to predict the fracture bounding boxes, all metrics, especially recall and thus F1-Score decrease.
This decrease cannot be recovered by the FP-reduction.
Moreover, when comparing their results with its corresponding counterpart without, the FP-reduction hurts the performance across all metrics.
 
\begin{table}[t]
    \centering
    \caption{Quantitative results across different methods. Multi-label on full-image and multi-class on GT BBox have been trained on GT fracture locations and AO codes. YOLO BBox describes the patched-based multiclass approach using YOLO for fracture detection with given confidence (Conf) and optional FP-Reduction.}
    \begin{tabular*}{\textwidth}{@{\extracolsep\fill}lccccc}\hline
        \textbf{Model} & \textbf{FP-Reduction} &\textbf{Accuracy} & \textbf{F1-Score} & \textbf{Precision} & \textbf{Recall} \\\hline
        Multi-label on full-image& $\times$ & \textbf{0.8790} & 0.6771 &  0.6941 &  0.6631\\
        Multi-class on GT BBox & $\times$ & 0.7334 & \textbf{0.7630} &  \textbf{0.8085} & \textbf{0.7334} \\\hline
        \multirow{2}{*}{YOLO BBox, Conf 0.01} & $\times$ & 0.6603 & 0.2786 & 0.7655 & 0.1761 \\
        & \checkmark & 0.6593& 0.2460&  0.7351& 0.1545\\\hline
        \multirow{2}{*}{YOLO BBox, Conf 0.05} & $\times$ & 0.6463& 0.2421& 0.6872&  0.1516\\
        & \checkmark &0.6415& 0.2343& 0.6893 &  0.1451\\\hline
        \multirow{2}{*}{YOLO BBox, Conf 0.5} & $\times$ & 0.6171&  0.2126& 0.7497& 0.1287\\
        & \checkmark & 0.6171&  0.2126& 0.7497 & 0.1287 \\\hline
        \multirow{2}{*}{YOLO BBox, Conf 0.85} & $\times$ & 0.4775& 0.1100&  0.8673& 0.0607\\
        & \checkmark & 0.4751& 0.1099&  0.7763 &  0.0603 \\\hline
    \end{tabular*}
    \label{tab:comparison method}
\end{table}

\begin{SCfigure}
    \centering
    \includegraphics[width=0.45\textwidth]{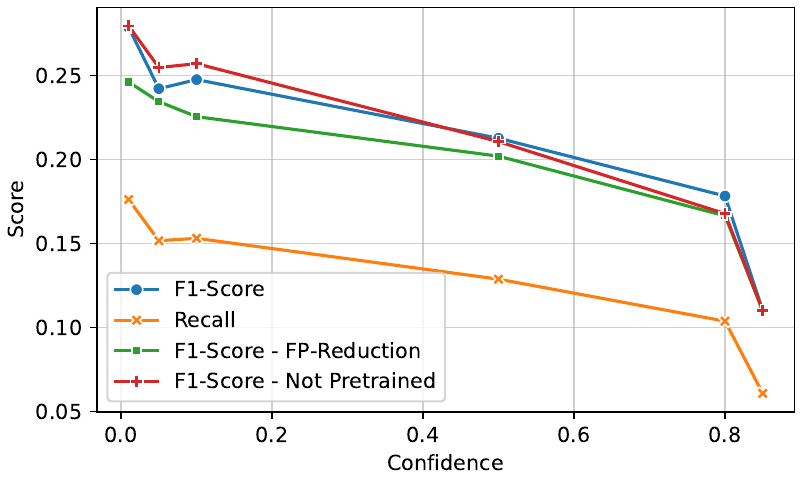}
    \caption{The impact of FP-reduction and pretraining on F1-scores was analyzed across different YOLO fracture detection confidence levels. The recall curve shows that lowering the confidence threshold increases recall.}
    \label{fig:confidence}
    \altText{Shows line plots that illustrate the metric values of various models in relation to the confidence threshold. The figure displays both the F1-score and the recall of the pretrained model, in addition to the F1-score of the false-positive reduction model and the non-pretrained model. It is evident that all lines demonstrate a decrease in their confidence scores, whilst concurrently exhibiting analogous shapes. The F1-Score of the pretrained model, as well as the non-pretrained and false-positive reduction model, are found to be highly comparable while the F1-Score of the pretrained model is most stable over all confidence scores. It is evident that the recall possesses an identical shape; however, its score is reduced by approximately 0.1 for each confidence score.}
\end{SCfigure}

Fig.~\ref{fig:confidence} reveals that reducing the confidence level for the YOLO prediction indeed increases the recall (orange, x-dotted line) and F1-score (blue, circle-dotted line), indicating a boost in the overall performance.
As already observed in Tab.~\ref{tab:comparison method}, the FP-reduction does not increase the performance.
When considering the results for the non-pretrained model (red, cross-dotted line), we find that utilizing ImageNet weights play no crucial role in our setting.

\begin{figure}[b]
    \centering
    \includegraphics[width=1\linewidth]{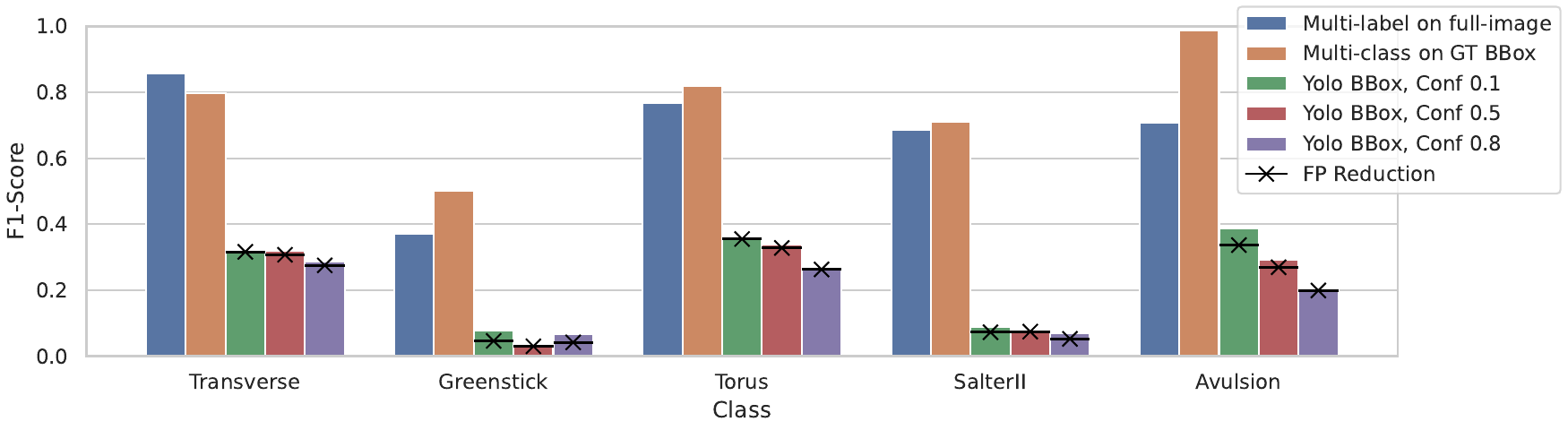}
    \caption{F1-scores for different fracture morphologies. See the caption of Tab.~\ref{tab:comparison method} for brief method descriptions.}
    \label{fig:f1perclass}
    \altText{The figure displays five bar plot groups for each morphology class one. In each group, the F1-scores of several models are displayed. These include the multi-label on full-image, multi-class on GT BBox and Yolo BBox from the confidence levels 0.1, 0.5 and 0.8. The F1-scores are accompanied by the corresponding FP-reduction values, indicated by a cross and a line. It is evident that the multi-class on GT BBox demonstrates superior performance in all classes, with the exception of the transverse class. The Yolo BBox model F1-Scores demonstrate a significant decrease in performance when compared to the GT-based models. It is evident that an increase in confidence in the Yolo approaches is associated with a decrease in the F1-score.}
\end{figure}

Fig.~\ref{fig:f1perclass} plots the F1-Scores for our five fracture morphologies across all models.
Except for “Transverse”, the patch-based approach utilizing the GT fracture bounding boxes yields the highest F1-score ($7.89\,\%$ on average) with the largest margin (0.7831) on the “Avulsion” class.
Again, it can be seen that the F1-Scores of the YOLO boxes on lower confidence levels are mostly higher in comparison to higher confidence levels and the FP-reduction does not boost the performances.

\section{Discussion}
Our proposed method to extract fracture morphology by assigning global AO codes to the corresponding fracture bounding boxes (Fig.~\ref{fig:pipeline} allow the reformulation of a global multi-label task to a local multi-class one.
Our study reveals, that this reformulation improves the overall performance ($7.89\,\%$ F1 on average), since it permits only one morphology to be present.
However, the real-world usability of the patch-based approach is currently limited by the unreliable YOLO fracture detection ($-65.3\,\%$ F1 at $0.85$ confidence level).
To overcome this, we investigated into multiple confidence thresholds for the fracture detection combined with a FP-reduction (extending the classifier with “Healthy” class) to handle the expected increase in false detection.
While we found a correlation in lowering the confidence level and the performance ($-48.44\,\%$ F1 at $0.01$), the FP-reduction does not yield any improvement.
We perform a sanity check by training a binary classifier (“Fracture” and “Healthy”) on the YOLO proposed patches with confidence $0.01$.
This yields a F1-score of $0.5$ (not shown in the results) indicating that a (preprocessing) step within our setup might limit the FP-reduction.
Hence, extending the fracture morphology classification task with the FP-reduction adds additional difficulty to the already complicated task.
The YOLO model frequently misses several bounding boxes (e.g. $2\,953$ at confidence level of $0.01$) that are present within the GT bounding boxes.
The addition of these false-negative examples in the evaluation metrics has led to a decrease in metric values, particularly in the recall.
While the performance with the GT bounding boxes proofs the potential of our multi-class reformulation, the focus of future work must be on the improvement of fracture detection algorithms to create a real-world application.
Furthermore, analysing a more diverse dataset, including more AO codes, would lead to a more generalized solution including more fracture morphologies.
 
\begin{acknowledgement}
	This research has been funded by Schleswig-Holstein, Grant Number 22023005.
    We thank Ludger Tüshaus and Franziska Halm from the University Hospital Schleswig-Holstein for their valuable expertise in extracting fracture morphology from AO codes.
\end{acknowledgement}

% This command generates the bibliography using the entries of the .bib file.
% Remove it only if you do not use a bibliography.
\printbibliography

\end{document}